\documentclass[10pt,twocolumn,letterpaper]{article}

\usepackage[pagenumbers]{cvpr} % To force page numbers, e.g. for an arXiv version

\usepackage{graphicx}
\usepackage{amsmath}
\usepackage{amssymb}
\usepackage{bm}
\usepackage{booktabs}
\usepackage{gensymb}
\usepackage[inline]{enumitem}
\usepackage{tabstackengine}
\usepackage{xcolor}

\usepackage[pagebackref,breaklinks,colorlinks]{hyperref}

\usepackage[capitalize]{cleveref}
\crefname{section}{Sec.}{Secs.}
\Crefname{section}{Section}{Sections}
\Crefname{table}{Table}{Tables}
\crefname{table}{Tab.}{Tabs.}

\def\cvprPaperID{3032} % *** Enter the CVPR Paper ID here
\def\confName{CVPR}
\def\confYear{2023}

\def\vec #1{\mathbf{#1}}	% vector
\def\mat #1{\mathbf{#1}}	% matrix
\def\cross #1{{}[#1]_{\times}}	% cross product operator
\def\mmatrx #1#2{\left[\begin{array}{#1} #2 \end{array}\right]}
\def\matrx #1{\begin{bmatrix} #1 \end{bmatrix}}

\def\oo {{\bm\omega}}	% twist vector

\def\Im {\mat I}		% identity matrix
\def\Zm {\mat 0}		% zero matrix
\def\Rm {\mat R}		% rotation matrix
\def\tv {\vec t}		% translation vector
\def\ox {\omega_x}
\def\oy {\omega_y}
\def\oz {\omega_z}
\def\oxa {\omega_{x_0}}
\def\oya {\omega_{y_0}}
\def\oza {\omega_{z_0}}
\def\oxb {\omega_{x_1}}
\def\oyb {\omega_{y_1}}
\def\ozb {\omega_{z_1}}

\def\doo {\Delta\oo}
\def\dox {\Delta\omega_x}
\def\doy {\Delta\omega_y}
\def\doz {\Delta\omega_z}

\def\dx {\Delta x}
\def\dy {\Delta y}
\def\df {\Delta f}

\begin{document}

\title{A Practical Stereo Depth System for Smart Glasses}
\author{
Jialiang Wang$^1$\thanks{Email: jialiangw@meta.com}~~~~~~~
Daniel Scharstein$^1$~~~~~~~
Akash Bapat$^1$~~~~~~~
Kevin Blackburn-Matzen$^2$\thanks{work was done at Meta}\\
Matthew Yu$^1$~~~~~
Jonathan Lehman$^1$~~~~~
Suhib Alsisan$^1$~~~~~
Yanghan Wang$^1$~~~~~
Sam Tsai$^1$\\
Jan-Michael Frahm$^1$~~~~
Zijian He$^1$~~~~
Peter Vajda$^1$~~~~
Michael F. Cohen$^1$~~~~
Matt Uyttendaele$^1$ \\[2mm]
$^1$Meta Platforms Inc.~~~~~~~
$^2$Adobe\\[-7mm]\mbox{~}
% {\tt\small jialiangw@meta.com}
}
\makeatletter
\let\@oldmaketitle\@maketitle
\renewcommand{\@maketitle}{\@oldmaketitle
  \centering\includegraphics[width=0.98\linewidth]{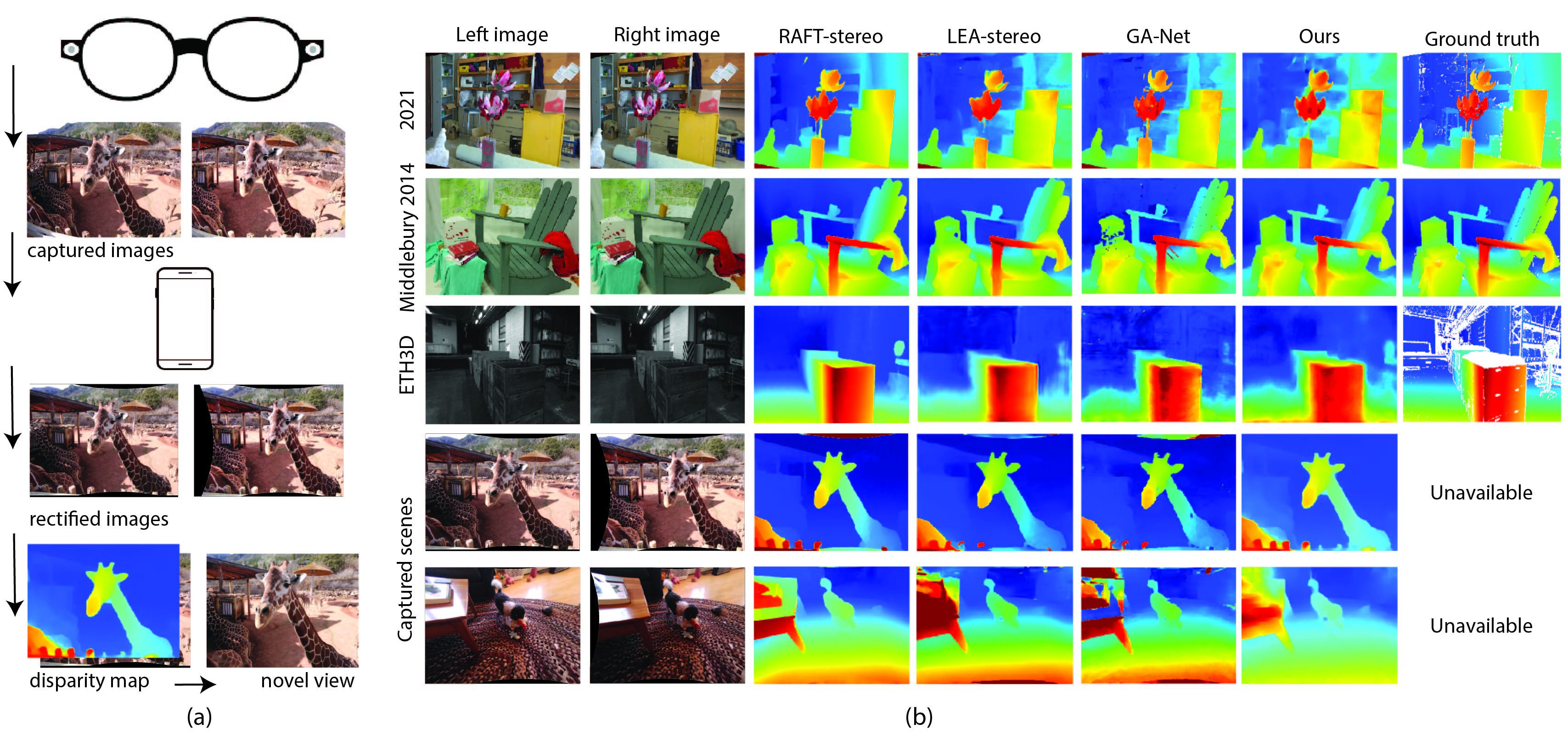}
  \vspace{-2mm}
  \captionof{figure}{\textbf{(a) Overview.} The user captures a stereo pair with smart glasses. The images are sent to the user's phone for processing, including online rectification and fast disparity prediction via neural networks. The predicted disparity map is then used to generate visual effects by rendering novel views. 
  \textbf{(b) Zero-shot accuracy.} Our stereo network, Argos, achieves high accuracy despite not being trained on these datasets, being quantized to 8 bits, and being significantly faster than SotA models such as RAFT-stereo~\cite{lipson2021raft}, GA-Net~\cite{Zhang2019GANet} and LEA-Stereo~\cite{cheng2020hierarchical}. 
 }
  \vspace{5mm}
  \label{fig:teaser}}
  
\maketitle
\begin{abstract}
\vspace{-1mm}
We present the design of a productionized end-to-end stereo depth sensing system that does pre-processing, online stereo rectification, and stereo depth estimation with a fallback to monocular depth estimation when rectification is unreliable.
The output of our depth sensing system is then used in a novel view generation pipeline to create 3D computational photography effects using point-of-view images captured by smart glasses. 
All these steps are executed on-device on the stringent compute budget of a mobile phone, and because we expect the users can use a wide range of smartphones, our design needs to be general and cannot be dependent on a particular hardware or ML accelerator such as a smartphone GPU. 
Although each of these steps is well studied, a description of a practical system is still lacking. For such a system, all these steps need to work in tandem with one another and fallback gracefully on failures within the system or less than ideal input data.
We show how we handle unforeseen changes to calibration, e.g., due to heat, robustly support depth estimation in the wild, and still abide by the memory and latency constraints required for a smooth user experience. We show that our trained models are fast, and run in less than 1s on a six-year-old Samsung Galaxy S8 phone's CPU. Our models generalize well to unseen data and achieve good results on Middlebury and in-the-wild images captured from the smart glasses. 
\end{abstract}

%%%%%%%%% BODY TEXT
\section{Introduction}
\label{sec:intro}

Stereo disparity estimation is one of the fundamental problems in computer vision, and it has a wide variety of applications in many different fields, such as AR/VR, computational photography, robotics, and autonomous driving. Researchers have made significant progress in using neural networks to achieve high accuracy in benchmarks such as KITTI~\cite{Menze2015CVPR}, Middlebury~\cite{Scharstein-gcpr2014} and ETH3D~\cite{schoeps2017cvpr}.

However, there are many practical challenges in using stereo in an end-to-end depth sensing system.
We present a productionized system in this paper for smart glasses equipped with two front-facing stereo cameras. The smart glasses are paired with a mobile phone, so the main computation happens on the phone.
The end-to-end system does pre-processing, online stereo rectification, and stereo depth estimation. In the event that the rectification fails, we fallback on monocular depth estimation.
The output of the depth sensing system is then fed to a rendering pipeline to create three-dimensional computational photographic effects from a single user capture. To our knowledge there is limited existing literature discussing how to design such an end-to-end system running on a very limited computational budget.

The main goal of our system is to achieve the best user experience to create the 3D computational photography effects. We need to support any mainstream phone the user chooses to use and capture any type of surroundings. 
Therefore, the system needs to be robust and operate on a very limited computational budget.
Nevertheless, we show that our trained model achieves on-par performance with state-of-the-art (SotA) networks such as RAFT-stereo~\cite{lipson2021raft}, GA-Net~\cite{Zhang2019GANet}, LEA-stereo~\cite{cheng2020hierarchical} and MobileStereoNet on the task of zero-shot depth estimation on the Middlebury 2014 dataset~\cite{Scharstein-gcpr2014} despite our network being orders of magnitude faster than these methods. 

\noindent The main technical and system contributions are:
\begin{enumerate}[itemsep=-1mm]
    \item We describe an end-to-end stereo system with careful design choices and fallback plans. Our design strategies can be a baseline for  other similar depth systems.
    \item We introduce a novel online rectification algorithm that is fast and robust.
    \item We introduce a novel strategy to co-design a stereo network and a monocular depth network to make both networks' output format similar.
    \item We show that our quantized network achieves competitive accuracy on a tight compute budget.
\end{enumerate}
\section{Related work}
\label{sec:related}
\begin{figure*}[t]
    \centering
    \includegraphics[width=0.93\textwidth]{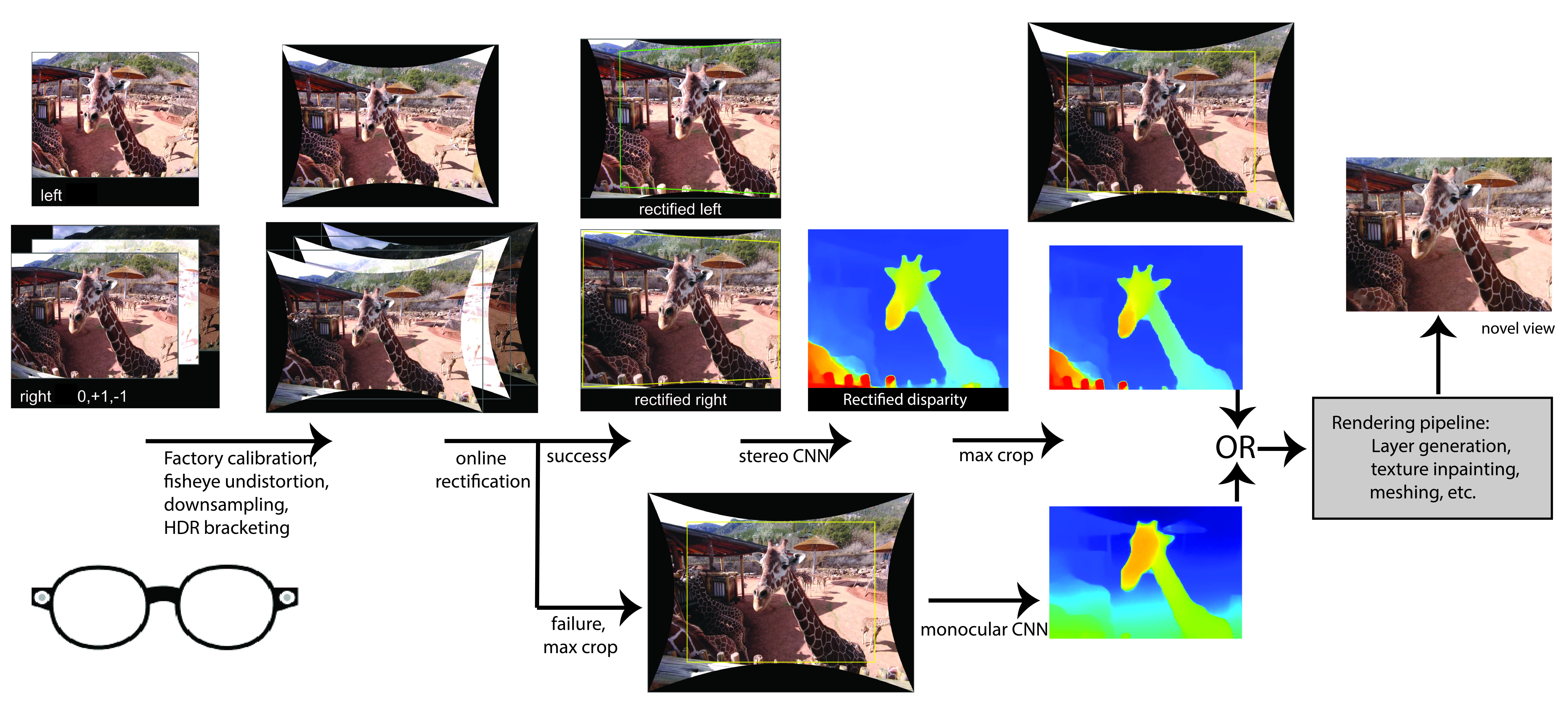}
    \caption{\textbf{Overview of the depth system.} The user captures a scene with smart glasses and saves the photo in their phone album. The HDR-bracketed right image is displayed in user's photo album. On the phone, the user can choose to make 3D effects from photos they took. When the user chooses to do so, we utilize the ``hidden'' left image to estimate stereo depth. The left-right image pair first goes through several pre-processing steps and then an online stereo rectification process to align the epipolar scanlines. We then either use a stereo matching network or a monocular depth estimation network depending on the status of rectification to obtain a relative disparity map. The cropped disparity map and the image are then used to render novel views to generate 3D photographic effects.
    \vspace{-2mm}
    }
    \label{fig:system_overview}
\end{figure*}

The focus of our work is building a complete stereo system for flexible-frame smart glasses with robust and light-weight processing on a mobile device.  This encompasses system design, online rectification, mono and stereo networks, and 3D effects.  While there is much related work for each component, few papers focus on the entire system design, despite its importance.
One example is the work by Kopf et al.~\cite{kopf2020one}, which presents an end-to-end system for one shot 3D photography, but the depth provider is a monocular depth network. Another example is the Google's work on Cinematic photos~\cite{Cinematic_Photos}, which also uses an encoder-decoder monocular depth estimation network. As for stereo, most recent deep-learning-based papers assume the input to the stereo depth system is calibrated and rectified~\cite{lipson2021raft, li2022practical, wang2019local, yang2019hsm, cheng2020hierarchical}. 
Barron et al.~\cite{Barron_2015_CVPR} proposed a bilateral approach for stereo to synthesize defocus that is used in Google's lens blur feature. 
Perhaps most relevant to our work is the work by Zhang et al.~\cite{zhang20202}, which uses dual cameras and dual pixels to estimate depth for defocus blur and 3D photos.
No paper, to the best of our knowledge, describes a stereo system with online rectification and failure handling when rectification fails.
Below we briefly summarize related work for each component in our design.

\noindent \textbf{Online stereo rectification:} 
Since imperfect rectification is a problem for almost all practical stereo systems,
a common approach is to allow stereo algorithm some vertical slack, 
e.g., by only utilizing horizontal gradients in the matching cost \cite{Hirschmuller-cvpr2009} 
or by training matching costs with samples that include small vertical disparities \cite{Zbontar-jmlr2016}.
Given that misalignment is a global property that can be modeled with just a handful of parameters, recovering these parameters
is clearly a cleaner approach.
Online stereo calibration and rectification has been researched
extensively 
\cite{Hartley-ijcv1999,Mallon-ivc2005,Dang-tip2009,Hansen-cvpr2012,Scharstein-gcpr2014,Ling-iros2016,Rehder-ivs2017,Kuhn-wacv2018,kumar_cvpr_ws_2018}.
Of specific interest are methods that robustly estimate rectification transforms for a single image pair from noisy feature correspondences \cite{Zilly-3dpvt2010,Georgiev-icip2013}.  Our formulation differs slightly and provides a family of models with increasing number parameters for describing the relative and absolute orientation of the two cameras.

\noindent \textbf{Stereo and monocular depth neural networks.}
A thorough review of the literature of stereo and monocular depth estimation neural networks is beyond the scope of this paper; see~\cite{Poggi-pami2022} for a recent survey. 

\noindent \textbf{Stereo.} The first category of papers is to use deep learning to learn the features used for stereo matching to replace handcrafted ones such as census~\cite{zabih1994non}. One notable paper in this category is MC-CNN~\cite{Zbontar-jmlr2016}. However, in recent years, the community has shifted to study end-to-end deep learning stereo. There are three types of architectures: 2D CNNs, 3D CNNs, and RNNs. 2D CNNs, e.g.,  \cite{tankovich2021hitnet, yin2019hierarchical}, are typically faster than 3D CNNs, such as~\cite{chang2018pyramid, cheng2020hierarchical}. More recently, RNN methods, e.g., RAFT-stereo~\cite{lipson2021raft} and CREstereo~\cite{li2022practical} have proven to show state-of-the-art performance~\cite{Scharstein-gcpr2014}, but they remain impractical to run on device. More recent works~\cite{tankovich2021hitnet, shamsafar2022mobilestereonet, yu_aicit_2022} try to make stereo networks run faster on device. There has also been recent interest in stereo robustness~\cite{robustvisionchalleng, wang2020improving}.

\noindent \textbf{Monocular depth.} There are supervised and unsupervised (self-supervised) monocular depth estimation papers using neural networks. In the supervised setting, researchers typically train an encoder-decoder network~\cite{Ranftl2022, kopf2020one} or a vision transformer~\cite{Ranftl2021}. The output is typically relative depth/disparity. In the self-supervised setting, rectified stereo pairs are typically used with photometric consistency loss~\cite{godard2019digging}, but the photometric consistency depth cue is absent in regions of stereo occlusion. There is also more recent work, such as~\cite{wu2022toward}, that combines the strengths of supervised and unsupervised methods.

\section{System overview}
\label{sec:sys_overview}

The overall system is shown in Figure~\ref{fig:system_overview}. Our smart glasses are equipped with a pair of hardware-synchronized fisheye cameras with resolution $2592\times 1944$. 
Once the user captures a scene, the image pair is transferred to a mobile phone for further processing.
In our system, the right image is used as the reference view for stereo matching and is used to render novel views.

On the user's smartphone, we first apply factory calibration to undistort the fisheye images to rectilinear and run HDR bracketing. 
We then downsample the images by $2\times$ to reduce computation for the next steps. 

We then run our novel robust online calibration and rectification algorithm, described in detail in Section~\ref{sec:calib}.  Online calibration is needed since the glasses deform during normal use, and extrinsics can vary significantly, in particular as a function of the user's head size.  Extrinsic and intrinsic parameters also change over time, especially with heat.
Our algorithm estimates 3-5 orientation parameters as well as relative focal length correction from a set of precise feature correspondences.  
This allows accurate rectification of the images. Online calibration may become unreliable however, for instance when the left camera is obstructed, or due to an insufficient number of feature matches. In this case, we fall back on using only the right image and obtain depth with a monocular depth estimation network.

To enable identical downstream processing, it is essential that stereo and monocular networks are similar to each other in terms of architecture output format. Therefore, while a typical stereo network can output absolute disparities, here we output \emph{relative disparities} normalized to 0..1 with both networks. Relative disparity is sufficient for our purpose of creating depth-based computational photography effects. Additionally, we share the encoder and decoder of the stereo network and the monocular network, with the stereo network having additional cost volume layers. We also train both networks with shared datasets, where the stereo dataset is created synthetically by rendering the second view using the monocular dataset. We discuss the networks and datasets in detail in Sections~\ref{sec:depth} and~\ref{sec:eval}.

Once a disparity map is computed, we crop the maximal valid region to preserve the original aspect ratio. Finally, the predicted disparity for the right camera and the corresponding color image is passed to a rendering pipeline to create the final three-dimensional effect. 

\section{Online rectification}
\label{sec:calib}
Two-view stereo algorithms commonly assume a rectified image pair as input \cite{Scharstein-ijcv2002}. For fixed stereo rigs, rectification is possible from factory calibration.
Our smart glasses, however, are made from material that can undergo significant bending ($>10\degree$).
This is further exacerbated by the variety of user's head sizes and changes of focal length with temperature. This almost always results in imperfect calibration parameters for stereo.
Imperfect rectification can strongly affect performance, in particular in high-frequency regions
and for near-horizontal image features
\cite{Hirschmuller-cvpr2009,Scharstein-gcpr2014}.

\begin{figure}[t]
    \centering
    \includegraphics[width=0.48\textwidth]{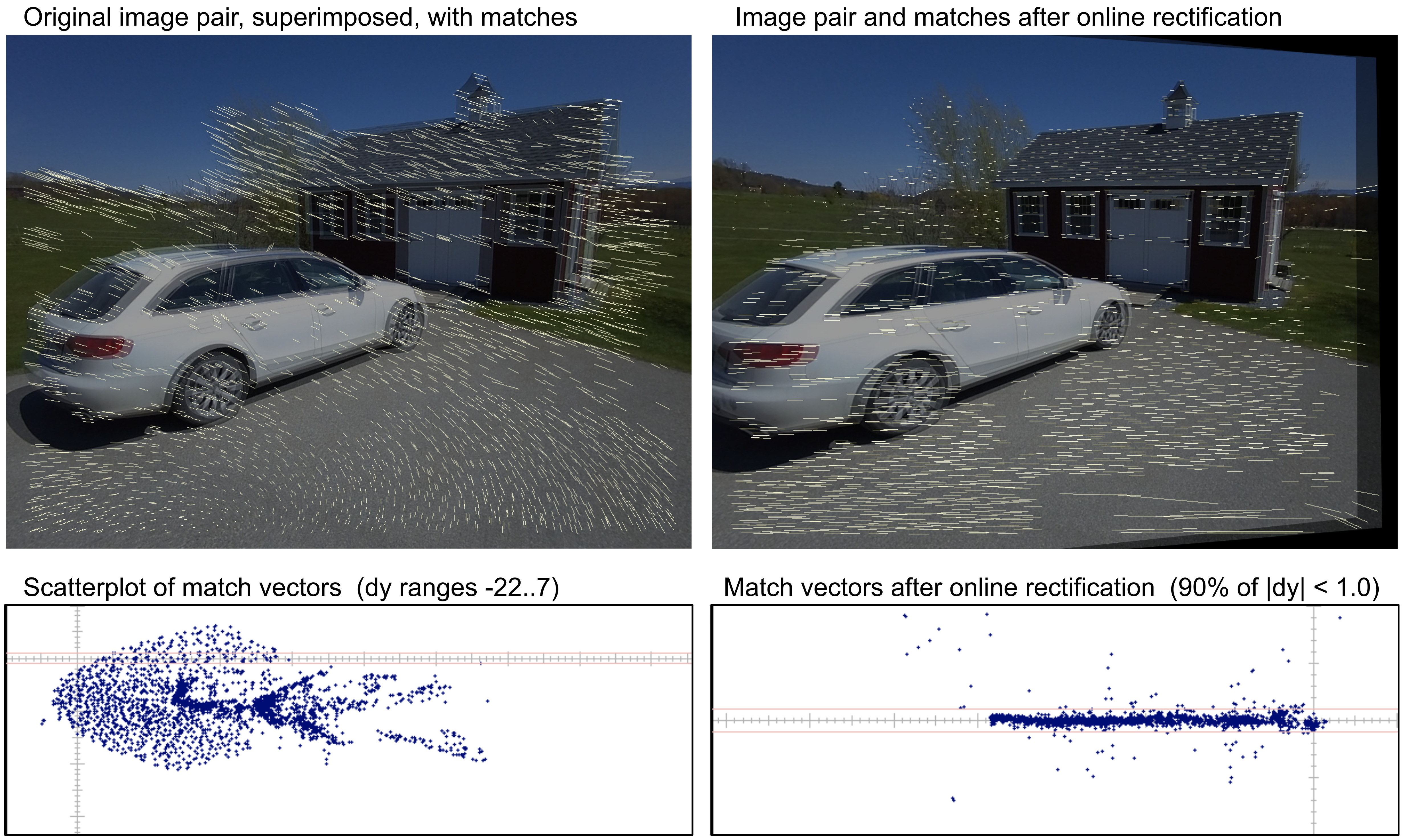}
    \caption{\textbf{Online rectification example.}
    Left: original image pair with matches overlaid, and scatterplot of match vectors, exhibiting large vertical disparities $\dy$.
    Right: image pair and matches after online rectification. More than 90\% of $|\dy|$ are below 1.0 pixels, and points at infinity are stabilized at $\dx=0$.}
    \label{fig:rectification}
\end{figure}

The goal of our online calibration method is thus to produce an accurately rectified stereo pair, given images with known lens distortion and estimates of camera intrinsics.  We do this from a set of precise feature correspondences, computed in the original fisheye images, and converted to rectilinear (pinhole) coordinates.
Figure~\ref{fig:rectification} shows the results of applying 
our method to a real image pair.

The basic idea is to model this rectification problem as the estimation of correcting rotations $\oo_0$, $\oo_1$, for each of the two cameras, keeping the baseline fixed.
This is a variant of the more common formulation of estimating extrinsics in the form of rotation $\Rm$ and translation $\tv$.
In both cases we need to estimate 5 parameters, since the absolute scale is unknown.
Existing formulations fix the length of the baseline $\|\tv\|$ and estimate its direction \cite{Scharstein-gcpr2014}
or simply keep $\tv_x$ constant \cite{Zilly-3dpvt2010,Georgiev-icip2013}.  
Here we adopt a simpler, symmetric approach --- we keep the world coordinate system aligned with the stereo rig,
both before and after re-rectification, define the length of the baseline to be $b = 1$,
and instead compute rotational corrections to both cameras.
We can hope to recover each camera's pan angles ($\oya$, $\oyb$) and roll angles ($\oza$, $\ozb$), 
but only relative pitch $\dox = \oxb - \oxa$ since absolute pitch (the angle of the rectifying
plane rotating about the x axis) is a free parameter.

In addition to these 5 parameters, which determine the extrinsics (relative pose), we
also estimate a 6th parameter $\df$ that models relative scale, i.e., a small change in focal length
$f_0 / f_1 = 1 + \df$. 
In practice we find that compensating for relative scale is very important, as focal lengths change with temperature.

\subsection{Projection model}
\label{sec:projection}

A 3D point $P = (X, Y, Z)$ projects to pixel coordinates $(u_0, v_0)$,  $(u_1, v_1)$ in the two
images.  Assuming that radial distortion has been corrected for and that 
intrinsics (focal lengths $f_i$ and principal points $(c_{x_i}, c_{y_i})$
are known, we convert to normalized image coordinates
%$(x_0, y_0)$, $(x_1, y_1)$ via
$ x_i = (u_i - c_{x_i}) / f_i$,
$ y_i = (v_i - c_{y_i}) / f_i$.

Under the above assumptions the cameras are located at
$\tv_0 = [0\;0\;0]^T$ and 
$\tv_1 = [1\;0\;0]^T$, and their rotations are $\Rm_0 = \Rm(\oo_0)$ and $\Rm_1 = \Rm(\oo_1)$.
We use a linear approximation for the rotations since we expect the rotational 
corrections to be small~\cite{Szeliski-book2020}:
$\Rm(\oo) \approx \Im + \cross{\oo}$.
In normalized image coordinates, point $P$ projects into the left camera at
\small{
\begin{equation}
[x_0 \; y_0 \; 1]^T \sim [\Im + \cross{\oo_0}] [X \; Y \; Z]^T .
%\matrx{x_0 \\ y_0 \\ 1} \sim
%\matrx{\\ \Im + \cross{\oo_0} \\ \; } \matrx{X \\ Y \\ Z} .
\end{equation}
}\normalsize
Parameterizing by inverse depth (\emph{disparity}) $d = 1/Z$ we can ``unproject'' the point
and project it into the right camera (see the supplementals for the derivation):
\small{
\setlength\arraycolsep{2pt}
\begin{eqnarray}
\matrx{x_1 \\ y_1 \\ 1} 
&\sim&
\matrx{\\ \Im + \cross{\doo}\\ \;}
\matrx{x_0 \\ y_0 \\ 1 } +
d \matrx{1 \\ -\ozb \\ \; \oyb} ,
\end{eqnarray}
}\normalsize
where $\doo = \oo_1 - \oo_0$ is the relative orientation of the two cameras.
We can see that for $d=0$ (a point at infinity), we can only recover the relative
orientation $\doo$.  For closer points ($d < 0$) we also get a constraint for
absolute roll and absolute pan, but not absolute pitch, as discussed earlier.

If we use $\dx = x_1 - x_0$ as our estimate for $d$ and also introduce a scale correction
$(1 + \df$), we can derive 
a constraint on $\dy = y_1 - y_0$
(see supplementals):
\small{
\setlength\arraycolsep{2pt}
\begin{equation}
\matrx{
 1\!+\!y_0 y_1 & - x_0 y_1 & -x_0 & -\dx y_1 & -\dx &  y_0}
\matrx{\dox \\ \doy \\ \doz \\ \oyb \\ \ozb \\ \df}
= \dy.
\label{eqn:fullsystem}
\end{equation}
}\normalsize

\subsection{Rectification algorithm}

We use Harris corners as feature candidates and match them across images using a hierarchical subpixel ZSSD feature matcher with left-right consistency check.
For each match we collect Eqn.~(\ref{eqn:fullsystem}) into an over-constrained system, which we solve using robust least squares, iterating with decreasing inlier thresholds for robustness to outliers. 

Furthermore, since our approximation of the true disparity $d$ via measured disparity $\dx$ only holds near the correct relative orientation, we first only solve for
$\doo$ and $\df$, whose equations do not depend on $d$:
\small{
\setlength\arraycolsep{3pt}
\begin{equation}
\matrx{1 + y_0\, y_1 & - x_0\, y_1 & -x_0 &  y_0}
\matrx{\dox \\ \doy \\ \doz \\ \df}
= \dy.
\label{eqn:halfsystem}
\end{equation}
}\normalsize
In practice, estimating these four parameters is very stable.  Once estimated,
we can then solve for one or both of the remaining parameters $\oyb$ and $\ozb$,
letting us recover absolute pan and roll for both cameras (since $\oya = \oyb\!-\!\doy$
and $\oza = \ozb\!-\!\doz$).
However, these absolute angles can only be estimated reliably if the observed
feature points are distributed over a sufficient depth range. 
In practice, we find that we usually get satisfactory rectification results for our low-resolution images with the 4-parameter model if we simply split the estimated relative angles evenly between left and right camera.

To declare online rectification successful, we require at least 100 matched feature points and an inlier rate of at least 60\% matches with $|\dy|\! \leq \! 1.0$.
In addition we check that the computed rectification angles stay within conservative bounds ($<\!5\degree$ absolute pitch and roll, and $<\!22\degree$ relative pan $|\doy|$).
If any of these criteria is not met, we fall back on monocular depth estimation.  In practice this happens for roughly 10\% of user-taken images, typically for feature-poor indoor scenes or if one of the cameras is obstructed.

We based our rectification design decisions on experiments
with 185 challenging test image pairs captured with prototypes, with manually established ground truth. Figure~\ref{fig:survey_results} shows a subset.
Table~\ref{tabledf} lists success and inlier rates for our 3 and 4-parameter models, and demonstrates that estimating $\df$ is important in practice.

\section{Co-design of monocular and stereo networks}
\label{sec:depth}
We describe a novel method to co-design stereo and monocular depth networks with the goal of having consistent output format (relative disparity), being light-weight, and being as accurate as possible. 

\subsection{Stereo network}

We design a stereo network with these components inspired by both classical and deep stereo methods~\cite{Szeliski-book2020}: 

    1) An encoder to extract multi-resolution features $f^l_0$ and feature $f^l_1$ independently from the input stereo pair, where $l = 1 \ldots L$, for $L$-level feature pyramids.
    
    2) Three-dimensional cost volumes that compare left and right feature distances
    % similar to the disparity space images built in stereo algorithms~\cite{chang2018pyramid, kendall2017end}.
    %We simply use 
    via 
    the cosine distance
    %between the left and right features 
    % rather than having the network to learn a feature similarity. 
    \begin{equation}
        C^l(x,y,d) = dot(f_0^l(x, y), f_1^l(x - d, y)),
        \label{eqn:cost_volume}
    \end{equation}
    for $d \in [0, \ldots, d_\text{max}]$. 
    
    3) A number of middle layers that take the cost volumes and the image features of the reference image as inputs, and aggregate the disparity information. Because the middle layers obtain information from cost volumes and reference images directly, they can better leverage monocular depth cues when stereo matching cues are weak (e.g., in textureless regions) or absent (e.g., in half-occluded regions).
    
\begin{figure}[t]
    \centering
    \includegraphics[width=0.48\textwidth]{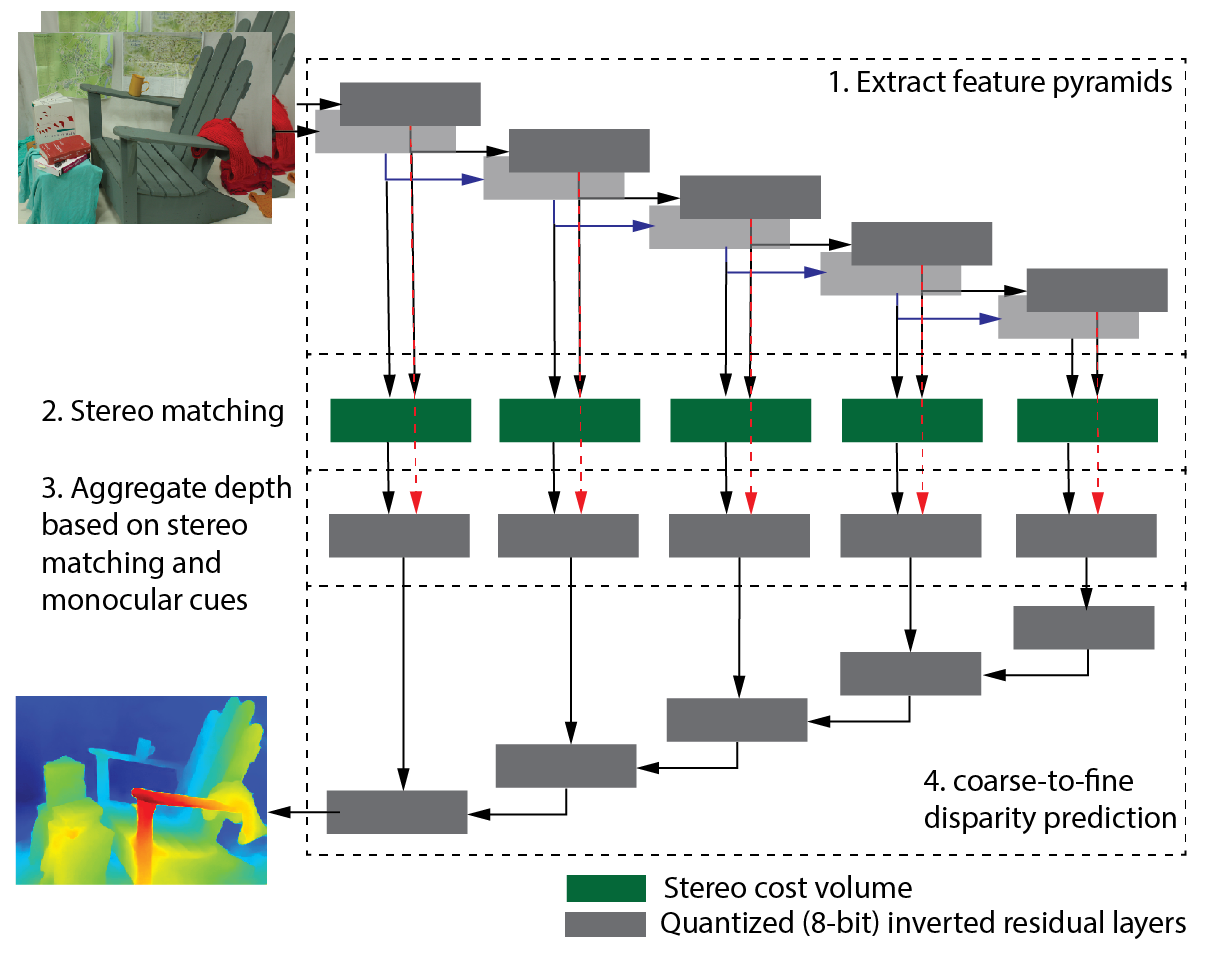}
    \caption{\textbf{Stereo network architecture.} We extract a feature pyramid for each input image, and then build 3D cost volumes. The 3D cost volumes and the features from the reference image are passed to middle layers to aggregate the disparity information. Finally, we design a coarse-to-fine decoder to compute a final disparity map. We use inverted residual blocks introduced in MobileNet v2~\cite{sandler2018mobilenetv2} and quantize to 8 bits to improve efficiency.
    }
    \label{fig:argos}
\end{figure}

    4) A coarse-to-fine decoder to predict an output disparity map. The output disparity map has the same resolution as the input right image. Each decoder module combines the output of the lower-resolution decoder module and the output of the same-resolution middle layers.
Figure~\ref{fig:argos} shows the architecture diagram of our stereo network.

\begin{table}[t]
{\footnotesize
\begin{tabular}{@{}l|l|l|l@{}}
Rectification model
&Success rate %$\uparrow$
&Median inlier rate %$\uparrow$
&Error $\df$  %$\downarrow$
% &$\doo$ error [deg\%]$\downarrow$
\\ \hline
3-param: $\doo$
&142/185= 77\%
&80\%  ~~(N=142)
&0.083\%
% &[0.55, 4.7,  1.07]
\\
4-param: $\doo$, $\df$
&{\bf 157}/185= {\bf 85\%}
&{\bf 85\%} ~(N=157) %, more difficult)
&\bf 0.024\%
% &\bf [0.39, 2.3, 0.56]
\end{tabular}
}
\vspace{-3mm}
\caption{Performance increase due to estimating relative scale $\df$.}
% \vspace{-1mm}
\label{tabledf}
\end{table}

\subsection{Monocular network}

\begin{figure}[t]
    \centering
    \includegraphics[width=0.48\textwidth]{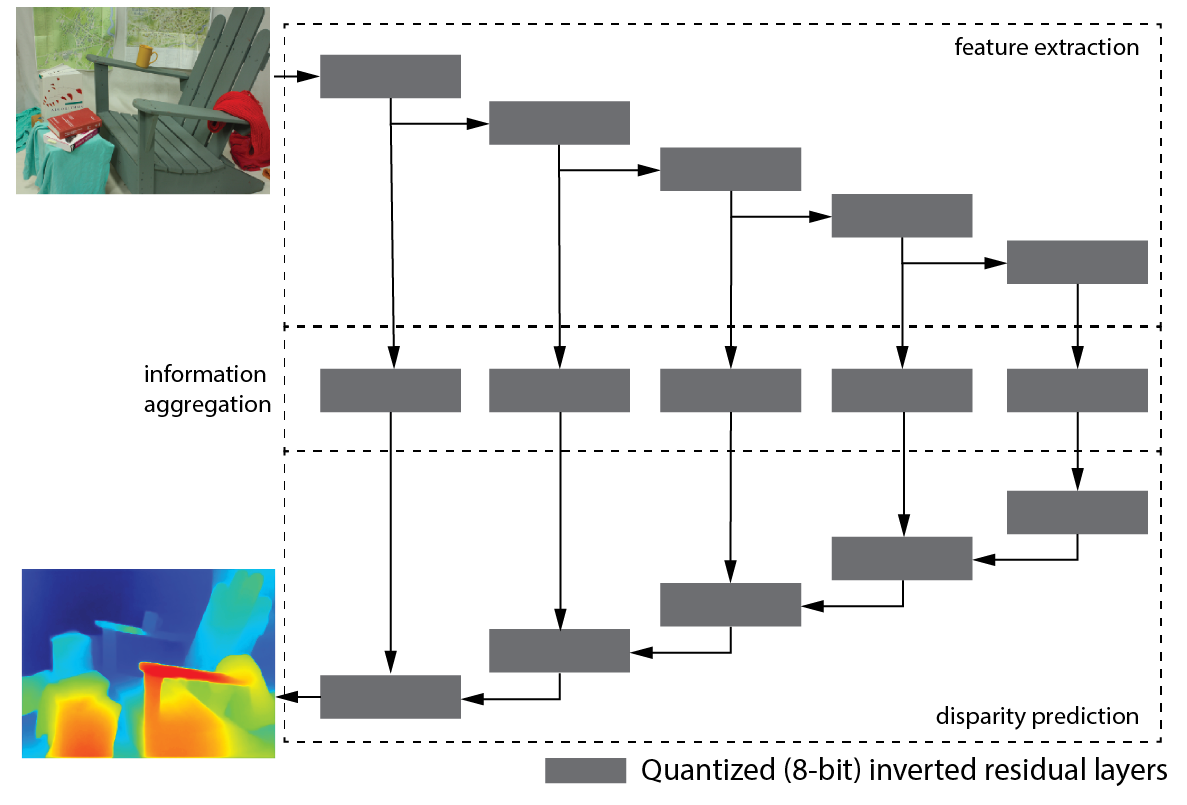}
    \caption{\textbf{Monocular depth network architecture.} We design an encoder-decoder architecture for our monocular depth estimation network. Similar to our stereo network, we use inverted residual blocks and 8-bit quantization to improve efficiency.}
    \label{fig:tiefenrausch}
\end{figure}

We design a monocular depth estimation network with three components: 
    1) An encoder to extract multi-resolution image features 
    $f^{l = 1 \ldots L}$;
    2) 
    %a few hourglass 
    middle layers to aggregate the depth information; and
    3) a coarse-to-fine decoder to predict a disparity map.
Figure~\ref{fig:tiefenrausch} shows the architecture diagram.

\subsection{Shared network components}
All three components of the monocular network also exist in the stereo network. The stereo network has additional cost volume modules to explicitly compare the similarity between the left and right local features. Therefore, we propose to share the exact same encoder,
%hourglass 
middle layers and decoder between the two networks. For stereo, the encoder runs twice to extract left and right image features independently (i.e., a Siamese encoder).

To train such networks with shared components, we train the monocular depth network first.
Then we initialize the stereo network with the trained weights from the monocular network, and fine-tune with stereo training data. Both networks are trained with the same loss
\begin{align}
     L(d_*, d_{gt}) & = \text{\it smooth\_L1}(d_*, d_{gt})  \nonumber \\
     & + \lambda \sum_{l=1}^L \text{\it smooth\_L1}(\text{\it grad}(d_*^l), \text{\it grad}(d_{gt}^l)),
\label{eqn:loss}
\end{align}
where
\begin{align*}
smooth\_L1(x, y) = \begin{cases}0.5\left(x - y\right)^2 , & \text { if }\left|x - y \right|< 1.0 \\ \left|x - y\right|-0.5, & \text { otherwise. }\end{cases}
\end{align*}

We use the median aligner proposed in MiDaS~\cite{Ranftl2022} to obtain aligned output disparity $d_*$ with ground truth $d_{gt}$. 
This design enables both the mono and stereo networks to output relative disparity maps and maximizes the consistency in predictions.

To improve efficiency, we use inverted residual modules proposed in MobileNetV2~\cite{sandler2018mobilenetv2} for all layers and quantize both weights and activation to 8 bits. 
Only the output layer and the layers that compute the cost volumes are not quantized. We keep the output layer 32 bits to get sub-pixel resolution and avoid quantization artifacts in the 3D effects.

\subsection{Novel training datasets}
% To train the networks on the same data, 
We derive the stereo data entirely from an internal 4M monocular dataset captured with iPhone.
Using the embedded depth map from the monocular dataset, we render the second view according to the known focal length and baseline of our smart glasses.
The occluded regions are also inpainted. We use the LDI-based rendering method discussed in~\cite{kopf2020one} which we will briefly discuss in Section~\ref{sec:novel_view}.
This makes our stereo dataset much more diverse than any existing training datasets.
Additionally, both the stereo and the monocular networks can be trained with the same data to make sure their output and behavior are similar, which is one of the main design goals of our depth sensing system. 

A similar technique was also described by Watson et al.~\cite{watson2020learning}, but they use a monocular depth estimation network such as MiDaS~\cite{Ranftl2022} to generate the depth.

To make the training data photo-realistic and challenging for the stereo network, we perform data augmentations for our rendered stereo dataset including brightness, contrast, hue, saturation, jpeg compression, reflection augmentation and image border augmentation. Reflection augmentation and image border augmentation are discussed in more details below. Other augmentations are standard.

\begin{figure*}[t]
    \centering
    \includegraphics[width=0.97\linewidth]{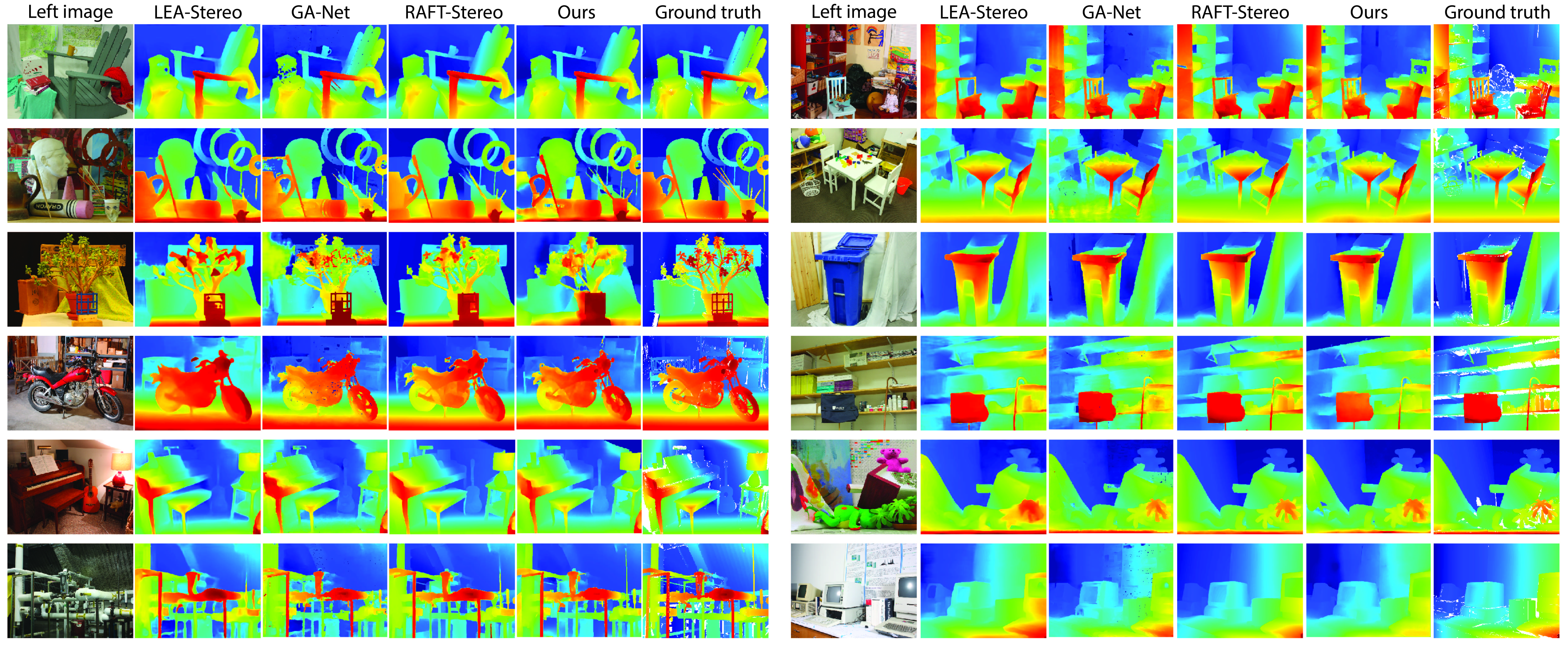}
    \caption{\textbf{Train on Sceneflow and test on Middlebury.} We achieve on par performance with SotA methods despite our model being much smaller and quantized.
    All methods are tested on $384 \times 288$ resolution. 
    }
    \label{fig:middlebury_results}
\end{figure*}

\begin{table*}
\small
\centering
\begin{tabular}{ p{3.6cm}|p{1.6cm}|p{1.4cm}|p{1.6cm}|p{1.2cm}|p{1.5cm}|p{1.6cm}|p{1.6cm} } 
\hline
   & Train data & Quantized & Latency (s) & AbsRel$\downarrow$ & $\delta>1.25$$\uparrow$ & $\delta>1.25^2$$\uparrow$ & $\delta>1.25^3$$\uparrow$ \\ 
 \hline
 RAFT-stereo~\cite{lipson2021raft} & Sceneflow & no & 224.75 & 0.075 & 0.923 & \textbf{0.973} & \textbf{0.987} \\ 
 \hline
 LEA-stereo~\cite{cheng2020hierarchical} & Sceneflow & no & 3.66 & 0.096 & 0.902 & 0.959 & 0.975 \\ 
 \hline
GA-Net~\cite{Zhang2019GANet} & Sceneflow & no & 7.73 & 0.128 & 0.893 & 0.944 & 0.961 \\ 
 \hline
{\small MobileStereoNet (2D)}~\cite{shamsafar2022mobilestereonet} & Sceneflow & no & 12.30 & 0.180 & 0.818 & 0.920 & 0.956 \\ 
 \hline
{\small MobileStereoNet (3D)}~\cite{shamsafar2022mobilestereonet} & Sceneflow & no & 18.96 & 0.139 & 0.876 & 0.933 & 0.963 \\ 
 \hline
  Argos (Ours) & Sceneflow & yes & 0.167 & 0.107 & 0.897 & 0.953 & 0.971 \\ 
 \hline
 \hline
  Argos (Ours) & internal & yes & 0.167 & \textbf{0.075} & \textbf{0.931} & 0.967 & 0.980 \\
 \hline
 \hline
  Tiefenrausch (mono)~\cite{kopf2020one} & internal & yes & 0.140 & 0.253 & 0.595 & 0.813 & 0.905 \\
 \hline
\end{tabular}
\caption{ 
% We use the pretrained RAFT-stereo, LEA-stereo and GA-Net weights provided on Github. 
\textbf{Middlebury results.} 
% Latency is reported on a Intel Xeon Gold 6138 CPU @ 2.00GHz, and tested on $384 \times 288$ resolution.  Our method performs on par with SotA methods when trained on Sceneflow, and overall best in several metrics when trained on our internal dataset; it is also significantly faster.
We report results on the 15 stereo pairs of the Middlebury 2014 training set. We use the pretrained PyTorch checkpoints in GitHub repos for baseline methods. For fair comparison, latency is measured at 384×288 resolution on a server without using GPU (Intel Xeon Gold 6138 CPU @ 2.00GHz).  Note that GPUs can yield significant speedup for some methods (e.g. 1.04s for RAFT-stereo on a Tesla V100 GPU). Torchscripting and quantization may also bring speedup for some comparison methods.}
\label{tab:middlebury_comparison}
\end{table*}

% \subsection{Augmentations}

% \textbf{Reflection augmentation.} 
Certain types of surfaces can reflect light, causing bright specular highlights in the images. This creates a difficult scenario for stereo matching because there are two textures at conflicting depths: one of the reflected light source and one of the reflecting surface.
Some existing work models reflections in stereo explicitly~\cite{tsin2003stereo}.  We instead train our stereo network to ignore reflections of specular highlights via simple augmentation.
We model a specular highlight as additive light following a Gaussian distribution with random  size $(\sigma_x, \sigma_y)$ and location $(\mu_x, \mu_y)$ in the left image.  We assume the light source is at random disparity $d_l$ and render its corresponding highlight in the right image at $(\mu_x - d_l, \mu_y)$.

% Daniel: no point showing these equations.
%Then the images after reflection augmentation are
%\begin{align*}
%    I_l^*(x,y) & = I_l(x, y) + \frac{1}{2 \pi \sigma_x \sigma_y} e^{\frac{-((x - c_w)^ 2}{(2 \sigma_x^2)} + \frac{(y - c_h)^2}{(2 \sigma_y^2))}} \\
%    I_r^*(x,y) & = I_r(x, y) + \frac{1}{2 \pi \sigma_x \sigma_y} e^{\frac{-((x - (c_w-d_l))^ 2}{(2 \sigma_x^2)} + \frac{(y - c_h)^2}{(2 \sigma_y^2))}}.
%\end{align*}

Finally, we also randomly add oval-shaped black image borders to mimic the invalid regions caused by warping and rectification in our depth pipeline (see Figure~\ref{fig:system_overview}).
\section{Novel view synthesis}
\label{sec:novel_view}
We use the same LDI (layered depth image) based method described in~\cite{kopf2020one} to create the stereo training dataset as well as to create our 3D effects.
Here we briefly review this method. Given a image/depth pair, we lift it to a layered depth image, hallucinate the occluded geometry by doing LDI inpainting, and finally convert it to a textured mesh.
For stereo dataset creation, we use the monocular ground-truth depth and the color image to create such a mesh and render the second view.
For the 3D effect, we use the prediction from our stereo system and a pre-defined trajectory to generate a smooth video of novel views. 

\section{Experiments}
\label{sec:eval}

\begin{figure*}[t]
    \centering
    \includegraphics[width=1.0\linewidth]{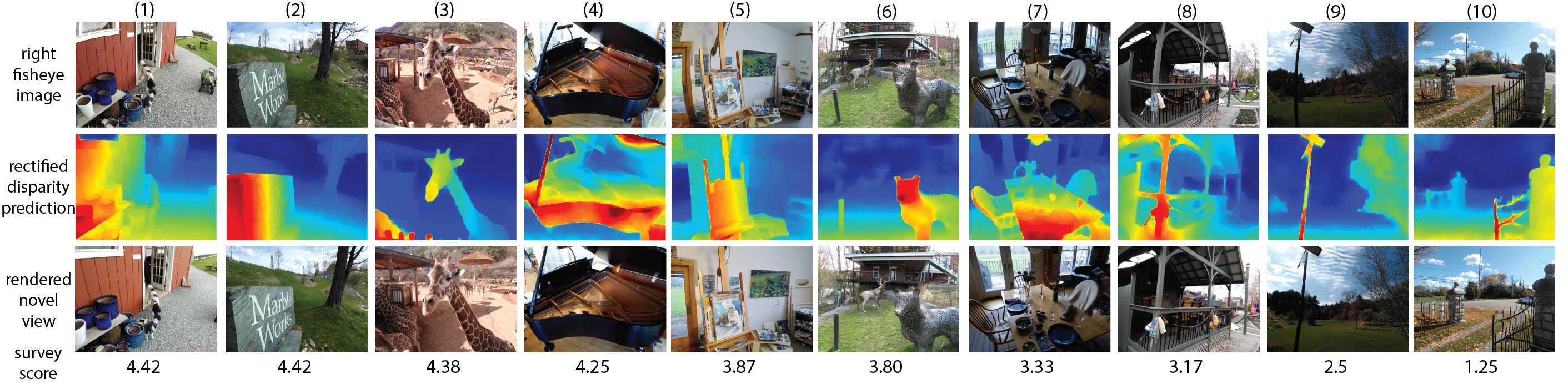}
    \vspace{-6mm}
    \caption{\textbf{Survey results.} 
    %We show 
    10 scenes captured from our smart glasses with original right images, rectified disparity maps, rendered novel views, and survey scores for rendered novel-view videos, with 1=poor and 5=excellent. Depth errors in thin structures often cause significant artifacts in rendered views. Depth errors in background or textureless regions typically do not lead to very noticeable artifacts.} 
    \label{fig:survey_results}
\end{figure*}

\noindent \textbf{Neural network implementation details.} 
We use the efficient monocular network \emph{Tiefenrausch} presented in~\cite{kopf2020one}. Details on network architecture can be found in the supplemental of~\cite{kopf2020one}. Our stereo network, \emph{Argos}, shares the same encoder, middle layers, and decoder with Tiefenrausch, but with added cost volume modules (Equation~\ref{eqn:cost_volume}), which are is parameter free. They simply take feature vectors of patches of the same epipolar scanline and compare patch similarity. We add one more layer after each cost volume module with input size $d_{\text{max}} + C(f_0^i)$ and output size $C(f_0^i)$, the dimension of the feature vector of the reference image.

For the production model, we first re-train  Tiefenrausch with our 4M internal iPhone dataset. As in~\cite{kopf2020one}, we align the predicted disparity map with the ``ground truth'' disparity map with the MiDaS median aligner~\cite{Ranftl2022} to output relative disparity. When computing the loss (Eqn.~\ref{eqn:loss}), we compute the gradient loss at $L = 5$ levels by simply doing $2 \times 2$ subsampling the prediction and the ground truth at coarser levels.
We then initialize the stereo network, Argos, with the Tiefenrausch weights. 
All but one layer (the added layer after cost volumes) can be initialized with the Tiefenrausch weights. We then finetune Argos with the rendered internal iPhone stereo dataset. During finetuning, we use Adam optimizer with base learning rate $4e-4$ and batch size of 24. We train a total of $640$k iterations, and use quantization-aware-training (QAT) in PyTorch~\cite{gholami2021survey, pytorch_qat}, which is essential to achieve high accuracy. PTQ (Post-training quantization) leads to significant accuracy drop. We start QAT at the $2000^{th}$ training iteration, and use the FBGEMM backend~\cite{fbgemm}. All input images are resized to $384 \times 288$ resolution for both training and evaluation, as this is the output disparity resolution used in our system.
To fairly compare our model with SotA models, we train another version of Argos with only the Sceneflow~\cite{MIFDB16} datasets (FlyingThings3D, Driving and Monkaa), and compare our results with other SotA models that are also trained on Sceneflow.

\vspace{1mm}
\noindent \textbf{Run time. }
We ran benchmarks of our pipeline on a Samsung Galaxy S8 CPU. The rectification pipeline takes 300-400ms, and the stereo network takes around 965ms. Other parts of the pipeline in total takes much lower latency than these two steps.
Our model is optimized for mobile CPU,
but converting SotA models to a mobile-friendly format is non-trivial and not very meaningful since they are not designed for mobile. As a compromise, we compare run time of all models on a computer server machine, with Intel(R) Xeon(R) Gold 6138 CPU @ 2.00GHz, shown in Table~\ref{tab:middlebury_comparison}.

\vspace{1mm}
\noindent \textbf{Middlebury Results.}
Table~\ref{tab:middlebury_comparison} and Figure~\ref{fig:middlebury_results} show the quantitative comparison of our method with several SotA stereo methods on the Middlebury 2014 dataset~\cite{Scharstein-gcpr2014}. When trained with Sceneflow dataset, our method achieves on-par performance with SotA methods despite being several orders of magnitude faster. Training with our rendered internal stereo dataset further improves our performance, and achieves the best Absolute Relative error among all tested models. 
We also include the results of Tiefenrausch~\cite{kopf2020one} retrained on our internal dataset. Comparing with Tiefenrausch, our stereo model, Argos, dramatically reduces absolute relative error from 0.253 to 0.075, which shows the effectiveness of our novel design to render a stereo dataset from a monocular dataset to train our model.

Note is that our goal is to design an end-to-end depth system that works robustly in all scenarios, rather than to achieve the best performance on academic benchmarks like  Middlebury~\cite{Scharstein-gcpr2014}. Some of our design choices, e.g., using reflection augmentation, can cause the accuracy to drop on Middlebury, which is carefully curated and do not have any scenes with obvious reflections. But such artifacts are common in real life. 
We are also one of the first to present a quantized 8-bit stereo model whereas all comparison models use 32-bit weights and activation. 
All of the above give us disadvantages.
Nevertheless, we achieve on-par performance with SotA methods, and are much faster.

\vspace{1mm}
\noindent \textbf{3D photo quality survey.}
To better understand the quality of our depth system, we conducted a survey with 24 participants. We capture a number of scenes from the smart glasses and then render novel view videos with depth maps from our depth system. We ask the participates to rate the quality of the rendered video from 1 to 5, with 5 being best (no artifacts) and 1 worst (major artifacts). Out of all responses, the average scores were 3.44 for using stereo and 2.96 for using mono depth. Figure~\ref{fig:survey_results} shows 10 survey scenes, covering both success and failure cases. One interesting observation is that the depth map quality sometimes does not directly correlate with the quality of the rendered novel view videos. For example, the depth map of scene (4) is poor, but there are few artifacts in the rendered view, and the users gave high scores. In contrast, the depth map of (10) is mostly correct except for the thin structures of the fence. However, the rendered video has jarring artifacts near the fence, causing poor user ratings in the survey. 
If we use standard stereo metrics such as the percentage of bad pixels, (10) likely has lower error than (4).
This highlights that solely comparing methods using standard metrics is not enough to evaluate a stereo method's performance in practice. Selected videos are attached in the supplement.

\section{Conclusion}
\label{sec:conclude}
We present the system design of an end-to-end stereo depth sensing system that runs efficiently on smartphones.
We describe a novel online rectification algorithm, a novel co-design of monocular and stereo disparity networks, and a novel method to derive large stereo datasets from monocular datasets.
We present an 8-bit quantized stereo model that has competitive performance on standard stereo benchmarks when compared against state-of-the-art methods.

\vspace{2mm}
\noindent \textbf{Acknowledgement: } We thank Rick Szeliski for his help in developing the online rectification model.

%%%%%%%%% REFERENCES
{\small
\bibliographystyle{ieee_fullname}
\bibliography{references}
}
\clearpage
\section{Appendix}

\subsection{Derivation of projection model}

Below is an expanded version of Section 4.1 with the full derivation of the projection model.

A 3D point $P = (X, Y, Z)$ projects to pixel coordinates $(u_0, v_0)$,  $(u_1, v_1)$ in the two
images.  Assuming that radial distortion has been corrected for and that 
intrinsics (focal lengths $f_i$ and principal points $c_{x_i}, c_{y_i})$
are known, we convert to normalized image coordinates
%$(x_0, y_0)$, $(x_1, y_1)$ via
$ x_i = (u_i - c_{x_i}) / f_i$,
$ y_i = (v_i - c_{y_i}) / f_i$.

Under the above assumptions the cameras are located at
$\tv_0 = [0\;0\;0]^T$ and 
$\tv_1 = [1\;0\;0]^T$, and their rotations are $\Rm_0 = \Rm(\oo_0)$ and $\Rm_1 = \Rm(\oo_1)$.
We use a linear approximation for the rotations since we expect the rotational 
corrections to be small:
\begin{equation}
\Rm(\oo) \approx
\Im + \cross{\oo} =
\matrx{ 1 & -\oz & \oy \\
        \oz & 1 & -\ox \\
       -\oy & \ox & 1}.
\end{equation}
In normalized image coordinates, point $P$ projects into the left camera at
\begin{equation}
\matrx{x_0 \\ y_0 \\ 1} \sim
\matrx{\\ \Im + \cross{\oo_0} \\ \; } \matrx{X \\ Y \\ Z} .
\end{equation}
Parametrizing by inverse depth (\emph{disparity}) $d = 1/Z$ we can ``unproject'' the point
\begin{equation}
\matrx{X \\ Y \\ Z \\ 1} \sim
\matrx{X/Z \\ Y/Z \\ 1 \\ 1/Z} \sim 
\mmatrx{c|c}{\\ \Im - \cross{\oo_0} & \Zm \\ \\ \hline \Zm & 1}
\matrx{x_0 \\ y_0 \\ 1 \\ d} .
\end{equation}
Projecting it into the right camera, we get
\medmuskip=0mu % horizontally compress equation
\thinmuskip=0mu
\thickmuskip=0mu
\small{
\setlength\arraycolsep{2pt}
\begin{eqnarray}
\matrx{x_1 \\ y_1 \\ 1} 
&\sim&
\mmatrx{c|c}{\\ \Im + \cross{\oo_1} & \Zm \\ \;}
\mmatrx{c|c}{&1\\ \;\;\;\; \Im \;\;\;\; & 0 \\ &0 \\ \hline \Zm & 1}
\mmatrx{c|c}{\\ \Im - \cross{\oo_0} & \Zm \\ \\ \hline \Zm & 1}
\matrx{x_0 \\ y_0 \\ 1 \\ d} \\ 
& \approx &
\matrx{\\ \Im + \cross{\oo_1 - \oo_0}\\ \;}
\matrx{x_0 \\ y_0 \\ 1 } +
\matrx{\\ \Im + \cross{\oo_1} \\ \; } \matrx{d \\ 0 \\ 0} \\
& = &
\matrx{\\ \Im + \cross{\doo}\\ \;}
\matrx{x_0 \\ y_0 \\ 1 } +
d \matrx{1 \\ -\ozb \\ \; \oyb} ,
\end{eqnarray}
}
\normalsize
\medmuskip=4mu % set back to normal
\thinmuskip=3mu
\thickmuskip=5mu
where $\doo = \oo_1 - \oo_0$ is the relative orientation of the two cameras.
We can see that for $d=0$ (i.e., a point at infinity), we can only recover the relative
orientation $\doo$.  For closer points ($d < 0$) we also get a constraint for
absolute roll and absolute pan, but not absolute pitch, as discussed earlier.

If we use $\dx = x_1 - x_0$ as our estimate for $d$ and also introduce a scale correction
$(1 + \df$), we have
\medmuskip=0mu % horizontally compress equation
\thinmuskip=0mu
\thickmuskip=0mu
\small{
\setlength\arraycolsep{1pt}
\begin{eqnarray}
\matrx{x_1 \\ y_1 \\ 1 + \df} 
&\sim&
\matrx{
1 & \doz & -\doy \\
-\doz & 1 & \dox \\
\doy & -\dox & 1}
\matrx{x_0 \\ y_0 \\ 1}
+
\dx \matrx{1 \\ -\ozb \\ \; \oyb} \\
&=&
\mmatrx{l}{
 x_0 + \doz\, y_0  - \doy \;\;+ \dx \\
 y_0 - \doz\, x_0  + \dox \;\;- \ozb \dx \\
\;1\;+ \doy\, x_0  - \dox\, y_0 + \oyb \dx }.
\end{eqnarray}
}
\normalsize
\medmuskip=3mu % horizontally compress equation
\thinmuskip=2mu
\thickmuskip=4mu

Cross-multiplying and dropping higher-order terms we get two equations in the 6 unknowns,
where $\dy = y_1 - y_0$:
\medmuskip=0mu % horizontally compress equation
\thinmuskip=0mu
\thickmuskip=0mu
\small{
\begin{equation}
% \setstacktabbedgap{0pt}
\setlength\arraycolsep{2pt}
\mmatrx{rrrrcr}{
   - y_0\, x_1 & 1 + x_0\, x_1 & -y_0 & \dx\, x_1 &    0 & -x_1 \\
 1 + y_0\, y_1 &   - x_0\, y_1 & -x_0 & -\dx\, y_1 & -\dx &  y_0}
\matrx{\dox \\ \doy \\ \doz \\ \oyb \\ \ozb \\ \df}
= \matrx{0 \\ \dy}.
%\label{eqn:fullsystem}
\end{equation}
}
\normalsize
\medmuskip=3mu % horizontally compress equation
\thinmuskip=2mu
\thickmuskip=4mu
Of these, only the second equation gives us a constraint relating $y_0$ and $y_1$.
We can collect these equations, one for each matched feature point, 
and solve the over-constrained system using robust least squares.

\end{document}